\documentclass[11pt,a4paper]{article}
\usepackage[hyperref]{emnlp2018}
\usepackage{times}
\usepackage{latexsym}

\usepackage{url}
\usepackage{amsmath}
\usepackage{graphicx}
\usepackage[lofdepth,lotdepth]{subfig}
\usepackage{multirow}
\usepackage[utf8]{inputenc}
\usepackage[T1]{fontenc}
\aclfinalcopy % Uncomment this line for the final submission
\def \mingyang {\textcolor{black}}

\title{A Visual Attention Grounding Neural Model\\for Multimodal Machine Translation}

\author{Mingyang Zhou \qquad
  Runxiang Cheng \qquad
  Yong Jae Lee \qquad
  Zhou Yu \\
  Department of Computer Science\\
  University of California, Davis\\
{\tt\{minzhou, samcheng, yongjaelee, joyu\}@ucdavis.edu}
  }
\date{}

\begin{document}
\maketitle
\begin{abstract}
%We introduce a novel neural structure, visual attention grounding, to utilize parallel vision information for machine translation. The visual attention grounding mechanism links the visual semantics in the image with the corresponding textual semantics. So the model jointly optimizes the learning of a visual-language shared embedding and translating between languages. 
We introduce a novel multimodal machine translation model that utilizes parallel visual and textual information. Our model jointly optimizes the learning of a shared visual-language embedding and a translator. The model leverages a visual attention grounding mechanism that links the visual semantics with the corresponding textual semantics. Our approach achieves competitive state-of-the-art results on the Multi30K and the Ambiguous COCO datasets. We also collected a new multilingual multimodal product description dataset to simulate a real-world international online shopping scenario.  On this dataset, our visual attention grounding model outperforms other methods by a large margin. %\yj{[YJ: but this isn't true for METEOR, right?]} %Because product descriptions are usually lengthy and our model is good at utilizing visual semantic attention to compensate the loss of distant textual history in sequence models.

\end{abstract}

\section{Introduction}

Multimodal machine translation is the problem of translating sentences paired with images into a different target language~\cite{ref1}. In this setting, translation is expected to be more accurate compared to purely text-based translation, as the visual context could help resolve ambiguous multi-sense words. Examples of real-world applications of multimodal (vision plus text) translation include translating multimedia news, web product information, and movie subtitles.  

Several previous endeavours \cite{ref13,ref9,ref16} have demonstrated improved translation quality when utilizing images. However, how to effectively integrate the visual information still remains a challenging problem. For instance, in the WMT 2017 multimodal machine translation challenge \cite{ref3}, methods that incorporated visual information did not outperform pure text-based approaches with a big margin.   

In this paper, we propose a new model called Visual Attention Grounding Neural Machine Translation (VAG-NMT) to leverage visual information more effectively. We train VAG-NMT with a multitask learning mechanism that simultaneously optimizes two objectives: (1) learning a translation model, and (2) constructing a vision-language joint semantic embedding. In this model, we develop a visual attention mechanism to learn an attention vector that values the words that have closer semantic relatedness with the visual context. The attention vector is then projected to the shared embedding space to initialize the translation decoder such that the source sentence words that are more related to the visual semantics have more influence during the decoding stage. When evaluated on the benchmark Multi30K and the Ambiguous COCO datasets, our VAG-NMT  model demonstrates competitive performance compared to existing state-of-the-art multimodal machine translation systems.

Another important challenge for multimodal machine translation is the lack of a large-scale, realistic dataset. To our knowledge, the only existing benchmark is Multi30K \cite{ref1}, which is based on an image captioning dataset, Flickr30K \cite{ref34} with manual German and French translations. There are roughly 30K parallel sentences, which is relatively small compared to text-only machine translation datasets that have millions of sentences such as WMT14 EN$\rightarrow$DE. Therefore, we propose a new multimodal machine translation dataset called IKEA to simulate the real-world problem of translating product descriptions from one language to another.  Our IKEA dataset is a collection of parallel English, French, and German product descriptions and images from IKEA's and UNIQLO's websites. We have included a total of 3,600 products so far and will include more in the future.  %Both the German and French descriptions on the websites are translated from English description according to our observations. However, sometimes therefore is still some noise, such as missing certain sentences in one language, we manually corrected such cases. Overall, this benchmark is a noisy but more realistic dataset that is a practical application of multimodal machine translation system. 

\section{Related Work}
%In the literature,there are two major streams to integrate vision information: approaches to employ separate attentions from different modalities and approaches to fuse vision information into NMT model as part of the input. The approach that uses separate attention mechanism will learn independent context vectors from a sequence of text encoder                                                                                                                       hidden states and a set of location-preserving vision features extracted from a pre-trained CNN network, where both attentions will affect the translation from decoder \cite{ref9,ref10}. The other approach instead extracts the global semantic feature and initialize either the encoder or decoder of the NMT to fuse the visual context.\cite{ref11,ref12}. Both methods demonstrate significant improvement over their Text-Only NMT baseline, but they are still worse than the best monomodal machine system in the WMT 2017 shared task workshop \cite{ref39}.

In the machine translation literature, there are two major streams for integrating visual information: approaches that (1) employ separate attention for different (text and vision) modalities, and (2) fuse visual information into the NMT model as part of the input. The first line of work learns independent context vectors from a sequence of text encoder hidden states and a set of location-preserving visual features extracted from a pre-trained convnet, and both sets of attentions affect the decoder's translation \cite{ref9,ref10}. The second line of work instead extracts a global semantic feature and initializes either the NMT encoder or decoder to fuse the visual context~\cite{ref11,ref12}. While both approaches demonstrate significant improvement over their Text-Only NMT baselines, they still perform worse than the best monomodal machine translation system from the WMT 2017 shared task~\cite{ref39}.%shared task 

%There are some other methods to employ the image context. \cite{ref13} embedded regional features extracted from a region-proposal network \cite{ref14} along with the word sequence as the input to encoder, which leads to significant improvement over their NMT baseline. The best multimodal machine translation system in WMT 2017 performed element-wise multiplication of the target language embeddings with an affine transformation of a CNN image feature vector as the mixed input in the decoder~\cite{ref15}. While this method outperforms all other methods in WMT 2017 shared task workshop, there is still a big room to explore in order to further expand the advantage over the monomodal translation system. %Although the improvement gained with their approach to involve vision knowledge is not obvious compared to their Text-Only comparison model when evaluated with automatic machine translation metrics, the improvement is enlarged with human evaluation metrics. 

The model that performs best in the multimodal machine translation task employed image context in a different way. \cite{ref13} combine region features extracted from a region-proposal network \cite{ref14} with the word sequence feature as the input to the encoder, which leads to significant improvement over their NMT baseline. The best multimodal machine translation system in WMT 2017~\cite{ref15} %\yj{[YJ: is there a citation for this approach?]} 
performs element-wise multiplication of the target language embedding with an affine transformation of the convnet image feature vector as the mixed input to the decoder. While this method outperforms all other methods in WMT 2017 shared task workshop, the advantage over the monomodal translation system is still minor.%\yj{While this method outperforms all other methods in WMT 2017 shared task workshop, there is still a big room to explore in order to further expand the advantage over the monomodal translation system. [YJ: is this trying to say that the improvement is minor compared to the text-only translation system?  If so, it would be better to say that more directly as "big room" is ambiguous.]}

%The visual context grounding process in our model is closely related to the previous work of multimodal shared space learning. \cite{ref26} proposed a neural language model to learn visual-semantic embedding space by optimizing a ranking cost objective, where distributed representation helps with the image caption generation. \cite{ref41} developed a method to densely align the different objects in the image with corresponding text captions in the shared space, which further improves the quality of the caption generated for images. Later on, the multimodal shared space learning work is extended to multimodal multilingual shared space learning which demonstrates further help on the task of image retrieval and neural machine translation.\cite{ref42,ref43}. Elliot \cite{ref16} gives the first attempt to integrate the ideas of multimodal shared space learning to help multimodal machine translation. He developed a multi-task architecture called ``imagination'' by sharing an encoder between a primary task of a classical encoder-decoder NMT and an auxiliary task of visual feature reconstruction.

The proposed visual context grounding process in our model is closely related to the literature on multimodal shared space learning. \cite{ref26} propose a neural language model to learn a visual-semantic embedding space by optimizing a ranking objective, where the distributed representation helps generate image captions. \cite{ref41} densely align different objects in the image with their corresponding text captions in the shared space, which further improves the quality of the generated caption. In later work, multimodal shared space learning was extended to multimodal multilingual shared space learning. %\yj{which demonstrates further help on the task of image retrieval and neural machine translation [YJ: this is not clear]}~\cite{ref42,ref43}. 
\cite{ref43} learn a multi-modal multilingual shared space through optimization of a modified pairwise contrastive function, where the extra multilingual signal in the shared space leads to improvements in image-sentence ranking and semantic textual similarity task. \cite{ref42} extend the work from \cite{ref43} by using the image as the pivot point to learn the multilingual multimodal shared space, which does not require large parallel corpora during training.  Finally, \cite{ref16} is the first to integrate the idea of multimodal shared space learning to help multimodal machine translation. Their multi-task architecture called ``imagination'' shares an encoder between a primary task of the classical encoder-decoder NMT and an auxiliary task of visual feature reconstruction.

Our VAG-NMT mechanism is inspired by \cite{ref16}, but has important differences. First, we modify the auxiliary task as a visual-text shared space learning objective instead of the simple image \mingyang{reconstruction} objective. Second, we create a visual-text attention mechanism that captures the words that share a strong semantic meaning with the image, where the grounded visual-context has more impact on the translation. We show that these enhancements lead to improvement in multimodal translation quality over \cite{ref16}.

%In the following section, we will introduce in details about our method. 

%Our VAG-NMT mechanism is an extension from Elliot's work. We optimize their method by modifying the auxiliary task as a visual-text shared space learning objective instead of just image retrieval objective. Additionally, we created an visual-text attention mechanism that captures the words sharing a strong semantic meaning connection with the image, where the grounded visual-context will have more impact on. In the following section, we will introduce in details about our method. 

\begin{figure}[h]
\centering
\includegraphics[width=\linewidth]{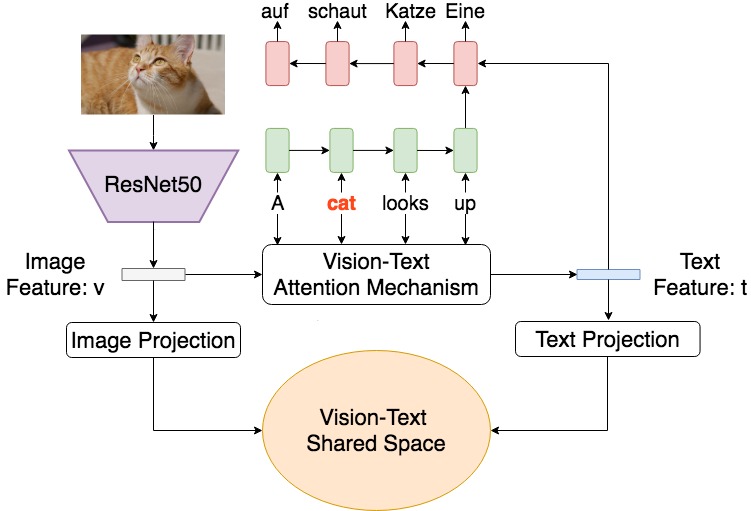}
\caption{An overview of the VAG-NMT structure}
\label{fig:figure1}
\end{figure}

\section{Visual Attention Grounding NMT}

Given a set of parallel sentences in language $X$ and $Y$, and a set of corresponding images $V$ paired with each sentence pair, the model aims to translate sentences $\{x_i\}_{i=1}^{N} \in X $ in language $X$ to sentences $\{y_i\}_{i=1}^{N} \in Y $ in language $Y$ with the assistance of images $\{v_i\}_{i=1}^{N} \in V$. 

We treat the problem of multimodal machine translation as a joint optimization of two tasks: (1) learning a robust translation model and (2) constructing a visual-language shared embedding that grounds the visual semantics with text. Figure~\ref{fig:figure1} shows an overview of our VAG-NMT model.  We adopt a state-of-the-art attention-based sequence-to-sequence structure \cite{ref2} for translation. For the joint embedding, we obtain the text representation using a weighted sum of hidden states from the encoder of the sequence-to-sequence model and we obtain the image representation from a pre-trained convnet. We learn the weights using a visual attention mechanism, which represents the semantic relatedness between the image and each word in the encoded text. We learn the shared space with a ranking loss and the translation model with a cross entropy loss. %The multimodal shared space is trained with a ranking loss, while the machine translation model is trained with a cross entropy loss. %that models the probability of the generated translation given the source text.  

\noindent The joint objective function is defined as:
\begin{equation}
    \begin{aligned}
    & J(\theta_{T},\phi_{V}) = \alpha J_T(\theta_{T}) +(1-\alpha)J_V(\phi_{V}),
    \end{aligned}
\end{equation}
where $J_T$ is the objective function for the sequence-to-sequence model, $J_V$ is the objective function for joint embedding learning, $\theta_{T}$ are the parameters in the translation model, and $\phi_{V}$ are the parameters for the shared vision-language embedding learning, and $\alpha$ determines the contribution of the machine translation loss versus the visual grounding loss.  Both $J_T$ and $J_V$ share the parameters of the encoder from the neural machine translation model. We describe details of the two objective functions in Section 3.2.

%We describe the design details of the two objective functions in the VAG-NMT model in the following subsections.

\subsection{Encoder}
We first encode an $n$-length source sentence $\{x\}$, as a sequence of tokens $x = \{x_1,x_2,\dots,x_n\}$, with a bidirectional \mingyang{GRU} \cite{ref25,ref46}. Each token is represented by a one-hot vector, which is then mapped into an embedding $e_i$ through a pre-trained embedding matrix. The bidirectional \mingyang{GRU} processes the embedding tokens in two directions: left-to-right (forward) and right-to-left (backward). At every time step, the encoder's \mingyang{GRU} cell generates two corresponding hidden state vectors: $\overrightarrow{h_i} = \overrightarrow{GRU(h_{i-1},e_i)}$ and $\overleftarrow{h_i} = \overleftarrow{GRU(h_{i-1},e_i)}$. The two hidden state vectors are then concatenated together to serve as the encoder hidden state vector of the source token at step $i$: $h_i = [\overleftarrow{h_i},\overrightarrow{h_i}]$.

\subsection{Shared embedding objective}%Visual attention grounding structure

After encoding the source sentence, we project both the image and text into the shared space to find a good distributed representation that can capture the semantic meaning across the two modalities. Previous work has shown that learning a multimodal representation is effective for grounding knowledge between two modalities \cite{ref26,ref27}. Therefore, we expect the shared encoder between the two objectives to facilitate the integration of the two modalities and positively influence translation during decoding. 

To project the image and the source sentence to a shared space, we obtain the visual embedding ($v$) from the pool5 layer of ResNet50 \cite{ref35} pre-trained on ImageNet classification~\cite{ref36}, and the source sentence embedding using the weighted sum of encoder hidden state vectors ($\{h_i\}$) to represent the entire source sentence ($t$). We project each $\{h_i\}$ to the shared space through an embedding layer. As different words in the source sentence will have different importance, we employ a visual-language attention mechanism---inspired by the attention mechanism applied in sequence-to-sequence models \cite{ref2}---to emphasize words that have the stronger semantic connection with the image. For example, the highlighted word ``cat" in the source sentence in Fig.~\ref{fig:figure1} has the more semantic connection with the image. %With such visual attention mechanism, we expect the visual context is grounded into those words having more representative power of visuals semantics.
% in semantic meanings

Specifically, we produce a set of weights $\beta = \{\beta_1,\beta_2,\dots,\beta_n\}$ with our visual-attention mechanism, where the attention weight $\beta_i$ for the $i$'th word is computed as:
\begin{equation}
    \beta_i = \frac{\exp(z_i)}{\sum_{l=1}^{N}\exp(z_{l})},
\end{equation}
and $z_i = \tanh(W_{v}v) \cdot \tanh(W_{h}h_i)$ is computed by taking the dot product between the transformed encoder hidden state vector $h_i$ and the transformed image feature vector $v$, and $W_{v}$ and $W_{h}$ are the association transformation parameters. 

Then, we can get a weighted sum of the encoder hidden state vectors $t = \sum_{i=1}^{n}\beta_i h_i$ to represent the semantic meaning of the entire source sentence. The next step is to project the source sentence feature vector $t$ and the image feature vector $v$ into the same shared space. The projected vector for text is: $t_{emb} = \tanh(W_{t_{emb}}t+b_{t_{emb}})$ and the projected vector for image is: $v_{emb} = \tanh(W_{v_{emb}}v+b_{v_{emb}})$.

We follow previous work on visual semantic embedding \cite{ref26} to minimize a pairwise ranking loss to learn the shared space:
\begin{equation}
\small{ 
\begin{aligned}
     J_V(\phi_V) &= \sum_p\sum_k \max\{0,\gamma-s(v_p,t_p)+s(v_p,t_{k \neq p})\} \\
    & + \sum_k\sum_p \max\{0,\gamma-s(t_k,v_k)+s(t_k,v_{p \neq k })\},
\end{aligned}
}
\end{equation}%gbp
where $\gamma$ is a margin, and $s$ is the cosine distance between two vectors in the shared space. \mingyang{$k$ and $p$ are the indexes of the images and text sentences, respectively.} $t_{k \neq p}$ and $v_{p \neq k }$ are the contrastive examples with respect to the selected image and the selected source text, respectively. When the loss decreases, the distance between a paired image and sentence will drop while the distance between an unpaired image and sentence will increase. 

%$s$ is a similarity metrics that computes the score of similarity between the two embeded vectors. In our work, we use cosine similarity. 

In addition to grounding the visual context into the shared encoder through the multimodal shared space learning, we also initialize the decoder with the learned attention vector $t$ such that the words that have more relatedness with the visual semantics will have more impact during the decoding (translation) stage. However, we may not want to solely rely on only a few most important words. Thus, to produce the initial hidden state of the decoder, we take a weighted average of the attention vector $t$ and the mean of encoder hidden states:
\begin{equation}
    s_0 = \tanh(W_{init}(\lambda t+(1-\lambda)\frac{1}{N}\sum_{i}^{N}h_i)),
\end{equation}
where $\lambda$ determines the contribution from each vector. Through our experiments, we find the best value for $\lambda$ is 0.5.
% and lose the information of other words during decoding stage

\subsection{Translation objective}%Decoder

During the decoding stage, at each time step $j$, the decoder generates a decoder hidden state $s_j$ from a conditional GRU cell \cite{ref8} whose input is the previously generated translation token $y_{j-1}$, the previous decoder hidden state $s_{j-1}$, and the context vector $c_j$ at the current time step:
\begin{equation}
    s_j = \text{cGRU}(s_{j-1},y_{j-1},c_j)
\end{equation}
The context vector $c_j$ is a weighted sum of the encoder hidden state vectors, and captures the relevant source words that the decoder should focus on when generating the current translated token $y_j$. The weight associated with each encoder hidden state is determined by a feed-forward network.
From the hidden state $s_j$ we can predict the conditional distribution of the next token $y_j$ with a fully-connected layer $W_o$ given the previous token's language embedding $e_{j-1}$, the current hidden state $d_j$ and the context vector for current step $c_j$:
\begin{equation}
     p(y_j | y_{j-1},x)= \text{softmax}(W_o o_t),
\end{equation}
where $o_t = \tanh(W_{e}e_{j-1}+W_{d}d_j+W_{c}c_j)$.  The three inputs are transformed with $W_e$, $W_d$, and $W_c$, respectively and then summed before being fed into the output layer. 

We train the translation objective by optimizing a cross entropy loss function:
\begin{equation}
     J_T(\theta_{T})= -\sum_{j}\log{p(y_j | y_{j-1},x)}
\end{equation}
By optimizing the objective of the translation and the multimodal shared space learning tasks jointly along with the visual-language attention mechanism, we can simultaneously learn a general mapping between the linguistic signals in two languages and grounding of relevant visual content in the text to improve the translation.  %through grounding the relevant visual context in text to

% \begin{equation}
%     c_j = \sum_{i=1}^{N} \alpha_{ji}h_i,
% \end{equation}
% where $\alpha_{ji}$ is the attention weight assigned to the encoder hidden state vector at step $i$.  It is computed by a feed-forward neural network with the input of the previous decoder hidden state $d_{j-1}$ and the encoder hidden state vector $v_a$:
% \begin{equation}
%     \alpha_{ji} = \frac{\exp(r_{ji})}{\sum_{j=1}^{N}\exp(r_{ji})},
% \end{equation}
% where $r_{ji} = v_a\tanh(W_a d_{j-1}+U_a h_i)$, and $W_a$ and $U_a$ are the learned parameters for the attention weights.The decoder predicts the conditional distribution of the next token $y_j$ 

\section{IKEA Dataset}

%We evaluated the proposed model on three datasets: Multi30K \cite{ref1}, Ambiguous COCO \cite{ref3} and our newly collected IKEA dataset.
%Multi30K dataset is the largest human-labeled dataset for multimodal machine translation. It consists of 31,014 images, where each image is annotated with an English caption and manual translations of image captions in German and French. There are 29,000 instances for training, 1,014 instances for validation, and 1,000 for testing. Additionally, we also evaluate our model on the Ambiguous COCO Dataset collected in the WMT2017 multimodal machine translation challenge \cite{ref3}. It contains 461 images and is a subset of the MSCOCO dataset \cite{ref37}. These images captions contain verbs with ambiguous meanings.% which are selected from VerSe dataset\cite{ref38}.

%Instead of focusing on solving real-world problems, previous multimodal machine translation data sets are all constructed based on image caption dataset  Previous machine translation datasets are constructed of short image captions, and  we propose a task of multimodal product description translation task that has a real-world impact on online shopping. 
\begin{table*}[h]
%\small
\centering
\begin{tabular}{lllll}
  & \multicolumn{2}{c}{English $\rightarrow$ German} & \multicolumn{2}{c}{English $\rightarrow$ French}\\
  \hline
  Method & BLEU & METEOR & BLEU & METEOR\\
  \hline
  Imagination \cite{ref16}     & $30.2$ & $51.2$ & N/A & N/A \\
  LIUMCVC \cite{ref15}    & $31.1 \pm  0.7$ & $52.2 \pm 0.4$ & $52.7 \pm 0.9$ & $69.5 \pm 0.7$ \\
  Text-Only NMT & $\boldsymbol{31.6 \pm 0.5}$ & $52.2 \pm 0.3$ & $53.5 \pm 0.7$ & $70.0 \pm 0.7$\\
  VAG-NMT & $\boldsymbol{31.6 \pm 0.3}$ & $52.2 \pm 0.3$ & $\boldsymbol{53.8 \pm  0.3}$ & $\boldsymbol{70.3 \pm 0.5}$ \\
\end{tabular}
\caption{Translation results on the Multi30K dataset}
\label{table:table_1}
\end{table*}

\begin{table*}[h]
%\small
\centering
\begin{tabular}{lllll}
   & \multicolumn{2}{c}{English $\rightarrow$ German} & \multicolumn{2}{c}{English $\rightarrow$ French}\\
   \hline
  Method & BLEU & METEOR & BLEU & METEOR\\
  \hline
  Imagination \cite{ref16} & $28.0$ & $\boldsymbol{48.1}$ & N/A & N/A \\
  LIUMCVC \cite{ref15}    & $27.1 \pm  0.9$ & $47.2 \pm 0.6$ & $43.5 \pm 1.2$ & $63.2 \pm 0.9$ \\
  \hline
  Text-Only NMT & $27.9 \pm 0.6$ & $47.8 \pm 0.6$ & $44.6 \pm 0.6$ & $64.2 \pm 0.5$\\
  VAG-NMT & $\boldsymbol{28.3 \pm 0.6}$ & $48.0 \pm 0.5$ & $\boldsymbol{45.0 \pm  0.4}$ & $\boldsymbol{64.7 \pm 0.4}$
\end{tabular}
\caption{Translation results on the Ambiguous COCO dataset}
\label{table:table_2}
\end{table*}

Previous available multimodal machine translation models are only tested on image caption datasets, we, therefore, propose a new dataset, IKEA, that has the real-world application of international online shopping. We create the dataset by crawling commercial products' descriptions and images from IKEA and UNIQLO websites. There are 3,600 products and we plan to expand the data in the future. Each sample is composed of the web-crawled English description of a product, an image of the product, and the web-crawled German or French description of the product. 

%%\yj{Since it is common for the descriptions of the same commercial product to be slightly different due to different regions}, 

Different than the image caption datasets, the German or French sentences in the IKEA dataset is not an exact parallel translation of its English sentence. Commercial descriptions in different languages can be vastly different in surface form but still keep the core semantic meaning. We think IKEA is a good data set to simulate real-world multimodal translation problems. The sentence in the IKEA dataset contains 60-70 words on average, which is five times longer than the average text length in Multi30K \cite{ref1}. These sentences not only describe the visual appearance of the product, but also the product usage. Therefore, capturing the connection between visual semantics and the text is more challenging on this dataset. The dataset statistics and an example of the IKEA dataset is in Appendix \ref{appendix:IKEA}. %The corpos level statistics of IKEA dataset is shown in Table \ref{table:table_0}. We also provide an EN$\rightarrow$DE example of our IKEA Dataset sample in Table \ref{table:table_5}. %We propose this dataset to help the multimodal machine translation research community to evaluate their models with a more practical scenario. 

\section{Experiments and Results}

\subsection{Datasets}
We evaluate our proposed model on three datasets: Multi30K \cite{ref1}, Ambiguous COCO \cite{ref3}, and our newly-collected IKEA dataset.  The Multi30K dataset is the largest existing human-labeled dataset for multimodal machine translation. It consists of 31,014 images, where each image is annotated with an English caption and manual translations of image captions in German and French. There are 29,000 instances for training, 1,014 instances for validation, and 1,000 for testing. Additionally, we also evaluate our model on the Ambiguous COCO Dataset collected in the WMT2017 multimodal machine translation challenge \cite{ref3}. It contains 461 images from the MSCOCO dataset \cite{ref37}, whose captions contain verbs with ambiguous meanings.% which are selected from VerSe dataset\cite{ref38}.

\subsection{Setting}
We pre-process all English, French, and German sentences by normalizing the punctuation, lower casing, and tokenizing with the Moses toolkit. A Byte-Pair-Encoding (BPE) \cite{ref44} operation with 10K merge operations is learned from the pre-processed data and then applied to segment words. We restore the original words by concatenating the subwords segmented by BPE in post-processing. During  training, we apply early stopping if there is no improvement in BLEU score on validation data for 10 validation steps. We apply beam search decoding to generate translation with beam size equal to 12. We evaluate the performance of all models using BLEU \cite{ref31} and METEOR \cite{ref30}. The setting used in IKEA dataset is the same as Multi30K, except that we lower the default batch size from 32 to 12; since IKEA dataset has long sentences and large variance in sentence length, we use smaller batches to make the training more stable. Full details of our hyperparameter choices can be found in Appendix \ref{appendix:Hyperparameters}. We run all models five times with different random seeds and report the mean and standard deviation. 
%to mitigate the variance in BLEU and METEOR

\subsection{Results}
\begin{table*}[h]
%\small
\centering
\begin{tabular}{lllll}
   & \multicolumn{2}{c}{English $\rightarrow$ German} & \multicolumn{2}{c}{English $\rightarrow$ French}\\
   \hline
  Method & BLEU & METEOR & BLEU & METEOR\\
  \hline
  %LIUMCVC-Mono    & $62.9\pm 3.6$ & $\boldsymbol{67.4\pm 0.2}$ & $65.1\pm 3.2$ & $\boldsymbol{71.4\pm 1.4}$  \\
  LIUMCVC-Multi    & $59.9\pm 1.9$ & $63.8\pm 0.4$ & $58.4\pm 1.6$ & $64.6\pm 1.8$ \\
  Text-Only NMT & $61.9 \pm 0.9$ & $65.6\pm 0.9$ & $65.2 \pm 0.7$ & $\boldsymbol{69.0 \pm 0.2}$\\
  VAG-NMT & $\boldsymbol{63.5 \pm 1.2}$ & $\boldsymbol{65.7\pm 0.1}$ & $\boldsymbol{65.8\pm 1.2}$ & \textbf{$68.9\pm 1.4$} \\
\end{tabular}
\caption{Translation results on the IKEA dataset}%Automatic Evaluation Metrics Results on IKEA German and IKEA French dataset}
\label{table:table_3}
\end{table*}

We compare the performance of our model against the state-of-the-art multimodal machine translation approaches and the text-only baseline. The idea of our model is inspired by the "Imagination" model \cite{ref16}, which unlike our models, simply averages the encoder hidden states for visual grounding learning. As "Imagination" does not report its performance on Multi30K 2017 and Ambiguous COCO in its original paper, we directly use their result reported in the WMT 2017 shared task as a comparison. LIUMCVC is the best multimodal machine translation model in WMT 2017 multimodal machine translation challenge and exploits visual information with several different methods. We always compare our VAG-NMT with the method that has been reported to have the best performance on each dataset.  

Our VAG-NMT surpasses the results of the ``Imagination" model and the LIUMCVC's model by a noticeable margin in terms of BLEU score on both the Multi30K dataset (Table~\ref{table:table_1}) and the Ambiguous COCO dataset (Table~\ref{table:table_2}). %, except the METEOR score for the German Ambiguous COCO Dataset. The
The METEOR score of our VAG-NMT is slightly worse than that of  "Imagination" for English -> German on Ambiguous COCO Dataset. This is likely because the ``Imagination" result was produced by ensembling the result of multiple runs, which typically leads to 1-2 higher BLEU and METEOR points compared to a single run.  Thus, we expect our VAG-NMT to outperform the ``Imagination" baseline if we also use an ensemble.

We observe that our multimodal VAG-NMT model has equal or slightly better result compared to the text-only neural machine translation model on the Multi30K dataset. On the Ambiguous COCO dataset, our VAG-NMT demonstrates clearer improvement over this text-only baseline. We suspect this is because Multi30K does not have many cases where images can help improve translation quality, as most of the image captions are short and simple. In contrast, Ambiguous COCO was purposely curated such that the verbs in the captions can have ambiguous meaning. Thus, visual context will play a more important role in Ambiguous COCO; namely, to help clarify the sense of the source text and guide the translation to select the correct word in the target language.

We then evaluate all models on the IKEA dataset. Table \ref{table:table_3} shows the results.
% \begin{table}[h!]
% \centering
% \begin{tabular}{|l|l|l|}
%   %\hline
%   \multicolumn{3}{c}{Multi30K Test 2017}\\
%   \hline
%   R@1 & R@5 & R@10 \\
%   \hline
%   64.0& 88.6 & 93.8 \\
%   \hline
% \end{tabular}
% \caption{Image retrieval Recall@K scores in Multi30K Dataset}
% \label{table:table_4}
% \end{table}
Our VAG-NMT has a higher value in BLEU and a comparable value in METEOR compared to the Text-only NMT baseline. Our VAG-NMT outperforms LIUMCVC's multimodal system by a large margin, which shows that the LIUMCVC's multimodal's good performance on Multi30K does not generalize to this real-world product dataset. The good performance may come from the visual attention mechanism that learns to focus on text segments that are related to the images. Such attention therefore teaches the decoder to apply the visual context to translate those words. This learned attention is especially useful for datasets with long sentences that have irrelevant text information with respect to the image.

\begin{figure}[ht]
\centering
\subfloat[Subfigure 1 list of figures text][\textit{a cyclist is wearing a helmet}]{
\includegraphics[width=0.45\textwidth]{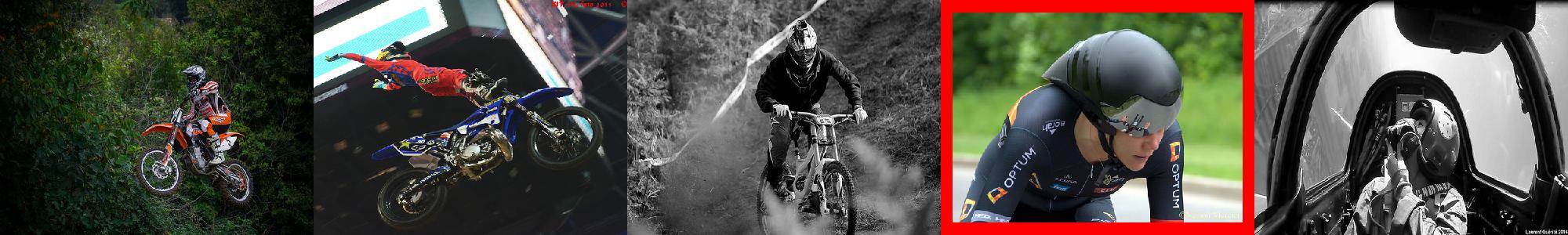}
\label{fig:subfig2}}
\qquad
\subfloat[Subfigure 2 list of figures text][ \textit{a black dog and his favorite toys.}]{
\includegraphics[width=0.45\textwidth]{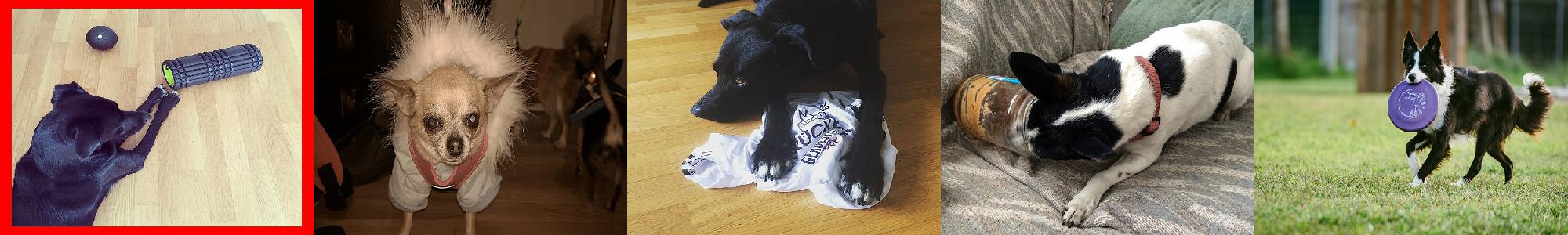}
\label{fig:subfig3}}
\caption{Top five images retrieved using the given caption. The original corresponding image of the caption is highlighted with a red bounding box.}
\label{fig:figure2}
\end{figure}

% \begin{table*}[h]
% \centering
% \begin{tabular}{lllll}
%               & MSCOCO      & Multi30K    & IKEA        & IKEA-CROSS  \\ \hline
% Text-Only-NMT  & 40          & \textbf{48} & 28          & NA          \\
% VAG-NMT       & \textbf{46} & 36          & \textbf{50} & \textbf{40} \\
% LIUMCVC Multi & NA          & NA          & NA          & 31          \\
% Tie           & 14          & 16          & 22          & 29  
% \end{tabular}

% \caption{Rater 2}%Automatic Evaluation Metrics Results on IKEA German and IKEA French dataset}
% \label{table:table_12}
% \end{table*}

\begin{figure*}[h]
\small
\centering
\begin{tabular}{p{5cm}p{5cm}p{5cm}}
\includegraphics[height=0.15\textwidth,width=5cm]{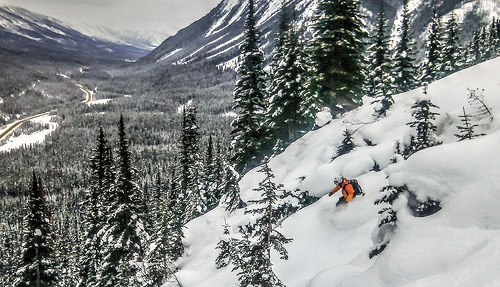} & \includegraphics[height=0.15\textwidth,width=5cm]{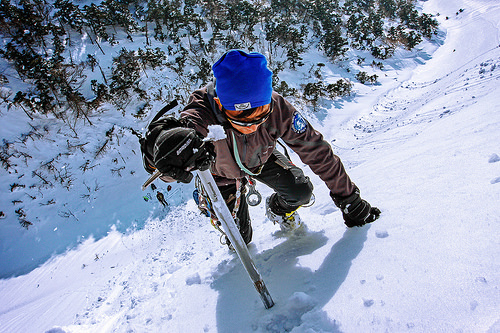} &
\includegraphics[height=0.15\textwidth,width=5cm]{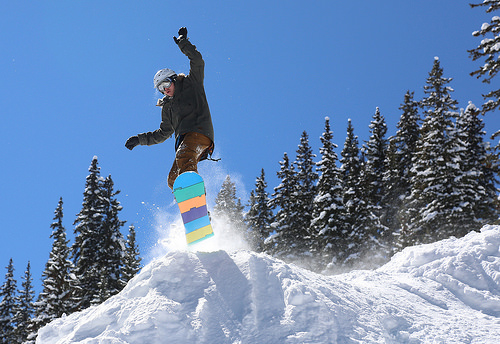}\\
a \textcolor{red}{person} is \textcolor{red}{skiing} or \textcolor{red}{snowboarding} down a mountainside . & 
a \textcolor{red}{mountain} climber trekking through the \textcolor{red}{snow} with a pick and a \textcolor{red}{blue} hat . &
\textcolor{red}{the snowboarder} is jumping in the \textcolor{red}{snow} .
\end{tabular}
\begin{tabular}{p{5cm}p{5cm}p{5cm}}
\includegraphics[height=0.15\textwidth,width=5cm]{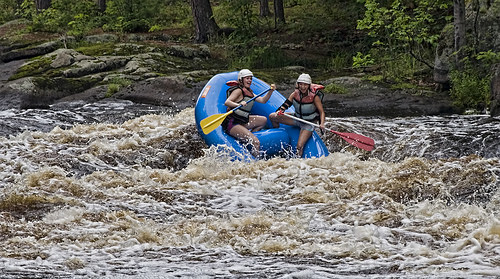} & \includegraphics[height=0.15\textwidth,width=5cm]{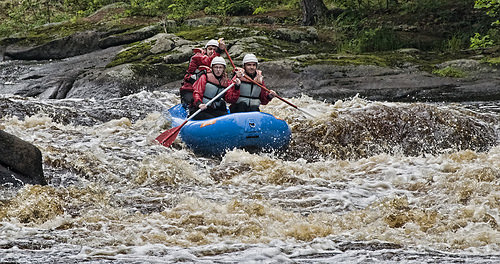} &
\includegraphics[height=0.15\textwidth,width=5cm]{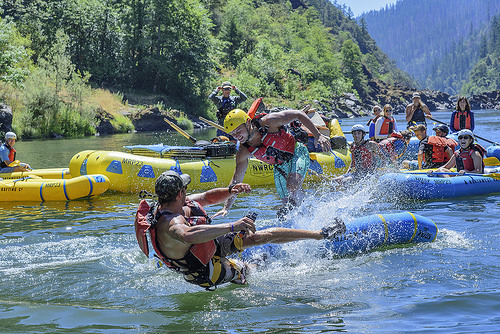}\\
two women \textcolor{red}{are} water \textcolor{red}{rafting .} & 
three people in a blue \textcolor{red}{raft} on a \textcolor{red}{river} of brown \textcolor{red}{water} . &
people in rafts watch as two men fall out of \textcolor{red}{their own rafts} .
\end{tabular}
\caption{\mingyang{Visual-text attention score on sample data from Multi30K.} The first and second rows show the three closest images to the caption \textit{a person is skiing or snowboarding down a mountainside} and \textit{two woman are water rafting}, respectively. The original caption is listed under each image. We highlight the three words with highest attention in red.}% for each caption%Automatic Evaluation Metrics Results on IKEA German and IKEA French dataset}
\label{table:attention_visualization}
\end{figure*}

% \begin{table*}[h]
% \small
% \centering
% \begin{tabular}{p{5cm}p{5cm}p{5cm}}
% \includegraphics[height=0.15\textwidth,width=5cm]{attention_visualization/1/37.jpg} & \includegraphics[height=0.15\textwidth,width=5cm]{attention_visualization/1/970.jpg} &
% \includegraphics[height=0.15\textwidth,width=5cm]{attention_visualization/1/79.jpg}\\
% two women \textcolor{red}{are} water \textcolor{red}{rafting .} & 
% three people in a blue \textcolor{red}{raft} on a \textcolor{red}{river} of brown \textcolor{red}{water} . &
% people in rafts watch as two men fall out of \textcolor{red}{their own rafts} .
% \end{tabular}

% \caption{The attention visualization for the three closest images to the caption \textit{two woman are water rafting .} The three words that has been assigned the highest attention weights are highlighted in red.}%Automatic Evaluation Metrics Results on IKEA German and IKEA French dataset}
% \label{table:attention_visualization_2}
% \end{table*}

\subsection{Multimodal embedding results}
To assess the learned joint embedding, we perform an image retrieval task evaluated using the Recall@K metric \cite{ref26} on the Multi30K dataset. 
%This metric represents the number of captions whose paired image is ranked within the top-K retrieved result. [Do not understand this sentence, please rewrite]

We project the image feature vectors $V = \{v_1, v_2,\dots, v_n\}$ and their corresponding captions $S = \{s_1,s_2,\dots,s_n\}$ into a shared space. We follow the experiments conducted by the previous visual semantic embedding work \cite{ref26}, where for each embedded text vector, we find the closest image vectors around it based on the cosine similarity. Then we can compute the recall rate of the paired image in the top K nearest neighbors, which is also known as $R@K$ score. %The image retrieval accuracy is shown in Table \ref{table:table_4}. 
The shared space learned with VAG-NMT achieves 64\% R@1, 88.6\% R@5, and 93.8\% R@10 on Multi30K, which demonstrates the good quality of the learned shared semantics. \mingyang{We also achieved 58.13\% R@1, 87.38\% R@5 and 93.74\% R@10 on IKEA dataset; 41.35\% R@1, 85.48\% R@5 and 92.56\% R@10 on COCO dataset.} Besides the quantitative results, we also show several qualitative results in Figure \ref{fig:figure2}. We show the top five images retrieved by the example captions. The images share "cyclist", "helmet", and "dog" mentioned in the caption.

\begin{figure*}[h]
\small
\centering
\begin{tabular}{c c p{9cm}}

\multirow{3}{*}{\includegraphics[height=0.1\textwidth,width=2.5cm]{Q1.jpg}} & Source Caption: & a tennis player is moving to the side and is gripping his \textcolor{red}{racquet} with both hands .\\[0.5 em]
& Text-Only NMT: & ein tennisspieler bewegt sich um die seite und greift mit beiden händen an \textcolor{red}{den boden} . \\[0.5 em]

& VAG-NMT: & ein tennisspieler bewegt sich zur seite und greift mit beiden händen \textcolor{red}{den schläger} . \\[0.5 em]

\multirow{3}{*}{\includegraphics[height=0.1\textwidth,width=2.5cm]{Q2.jpg}} & Source Caption: & three skiers skiing on a hill with two going down the hill and one moving \textcolor{red}{up} the hill .\\[0.5 em]
& Text-Only NMT: & drei skifahrer fahren auf skiern einen hügel hinunter und eine person fährt den hügel \textcolor{red}{hinunter} .\\[0.5 em]
& VAG-NMT: & drei skifahrer auf einem hügel fahren einen hügel hinunter und ein bewegt sich den hügel \textcolor{red}{hinauf} .\\ [0.5 em]

\multirow{3}{*}{\includegraphics[height=0.1\textwidth,width=2.5cm]{Q3.jpg}} & Source Caption: & a blue , yellow and green surfboard \textcolor{blue}{sticking out} of a sandy beach .\\[0.5 em]
& Text-Only NMT: & ein blau , gelb und grünes surfbrett \textcolor{blue}{streckt aus} einem sandstrand .\\[0.5 em]
& VAG-NMT: & ein blau , gelb und grüner surfbrett \textcolor{blue}{springt aus} einem sandstrand .\\[0.5 em]

\end{tabular}
\caption{Translations generated by VAG-NMT and Text-Only NMT. VAG-NMT performs better in the first two examples, while Text-Only NMT performs better in the third example. We highlight the words that distinguish the two systems' results in \textcolor{red}{red} and \textcolor{blue}{blue}. \mingyang{\textcolor{red}{Red} words are marked for better translation from VAG-NMT and \textcolor{blue}{blue} words are marked for better translation from Text-Only NMT. }} %\mingyang{The \textcolor{red}{red} word shows where VAG-NMT is better than Text-Only NMT. Vice Versa, the \textcolor{blue}{blue} word demonstrates where Text-Only NMT translate more accurately than VAG-NMT.}}
\label{table:table_6}
\end{figure*}

%%Ambiguous COCO dataset, Multi30K dataset and IKEA dataset. 
\subsection{Human evaluation}
We use Facebook to hire raters that speak both German and English to evaluate German translation quality. %All of the raters hired are students from University of California, Davis [Cannot mention, it is a conflict of anonymity]. 
%Due to the interest of time, we did not conduct evaluation on French translation results. 
As Text-Only NMT has the highest BLEU results among all baseline models, we compare the translation quality between the Text-Only and the VAG-NMT on all three datasets.
We randomly selected 100 examples for evaluation for each dataset. %We asked the user to choose between the two candidate translations we provided to them.  %Each evaluation sample is composed of a piece of source text in English, an image corresponds to the source text and the candidate translations from two models we want to compared. 
The raters are informed to focus more on semantic meaning than grammatical correctness when indicating the preference of the two translations. They are also given the option to choose a tie if they cannot decide. We hired two raters and the inter-annotator agreement is 0.82 in Cohen's Kappa. 

We summarize the results in Table \ref{table:table_11}, where we list the number of times that raters prefer one translation over another or think they are the same quality. Our VAG-NMT performs better than Text-Only NMT on MSCOCO and IKEA dataset, which correlates with the automatic evaluation metrics. However, the result of VAG-NMT is slightly worse than the Text-Only NMT on the Multi30K test dataset. This also correlates with the result of automatic evaluation metrics.

Finally, we also compare the translation quality between LIUMCVC multimodal and VAG-NMT on 100 randomly selected examples from the IKEA dataset. VAG-NMT performs better than LIUMCVC multimodal. The raters prefer our VAG-NMT in 91 cases, LIUMCVC multimodal in 68 cases, and cannot tell in 47 cases.

% We invited an experienced Natural Language Processing researcher as the third rater to evaluate data samples in which the previous two raters have different preferences on. Out of the 17 cases (we skipped the cases where one rater choose tie while the other choose a differently), 9 cases were tie, 4 ca

 %One possible reason is that there is irrelevant or biased information carried by the image feature vector which may introduces extra noise to the translation model. %We will investigate the hypothesis through further analysis of the quality of the two translation models.

\begin{table}[h]
\centering
\resizebox{0.48\textwidth}{!}
{
\begin{tabular}{llll}
         & MSCOCO      & Multi30K    & IKEA \\ \hline
Text-Only NMT & 76         & \textbf{72} & 75 \\
VAG-NMT       & \textbf{94} & 71 &  \textbf{82}   \\
Tie           & 30          & 57          & 43 
\end{tabular}
}
\caption{Human evaluation results }%Automatic Evaluation Metrics Results on IKEA German and IKEA French dataset}
\label{table:table_11}
\end{table}

\section{Discussion}

%\subsection{Does visual attention mechanism work?}
%We provide evidence and discuss how the attention mechanism ground the visual context into the sequence to sequence model to improve the translation quality. 
%As we have mentioned before, the attention mechanism is created to put emphasis on words that have a stronger connection with the visual semantics. 
To demonstrate the effectiveness of our visual attention mechanism, we show some qualitative examples in Figure \ref{table:attention_visualization}. Each row contains three images that share similar semantic meaning, which are retrieved by querying the image caption using our learned shared space. The original caption of each image is shown below each image. We highlight the three words in each caption that have the highest weights assigned by the visual attention mechanism.

In the first row of Figure~\ref{table:attention_visualization}, the attention mechanism assigns high weights to the words ``skiing", ``snowboarding", and ``snow". In the second row, it assigns high attention to ``rafting" or ``raft" for every caption of the three images. These examples demonstrate evidence that our attention mechanism learns to assign high weights to words that have corresponding visual semantics in the image. %is the shared semantic context among the three images, which demonstrate the success of capturing the words that have more semantic relatedness to the image.

We also find that our visual grounding attention captures the dependency between the words that have strong visual semantic relatedness. For example, in Figure~\ref{table:attention_visualization}, words, such as ``raft",``river", and ``water", with high attention appear in the image together. This shows that the visual dependence information is encoded into the weighted sum of attention vectors which is applied to initialize the translation decoder. When we apply the sequence-to-sequence model to translate a long sentence, the encoded visual dependence information strengthens the connection between the words with visual semantic relatedness. Such connections mitigate the problem of standard sequence-to-sequence models tending to forget distant history. This hypothesis is supported by the fact that our VAG-NMT outperforms all the other methods on the IKEA dataset which has long sentences.

Lastly, in Figure \ref{table:table_6} we provide some qualitative comparisons between the translations from VAG-NMT and Text-Only NMT. In the first example, our VAG-NMT properly translates the word "racquet" to ``den schläger", while the Text-Only NMT mistranslated it to ``den boden" which means ``ground" in English. We suspect the attention mechanism and visual shared space capture the visual dependence between the word ``tennis" and ``racquet". In the second example, our VAG-NMT model correctly translates the preposition ``up" to ``hinauf" but Text-Only NMT mistranslates it to ``hinunter" which means ``down" in English. We consistently observe that VAG-NMT translates prepositions better than Text-Only NMT. We think it is because the pre-trained convnet features captured the relative object position that leads to a better preposition choice. Finally, in the third example, we show a failure case where Text-Only NMT generates a better translation. Our VAG-NMT mistranslates the verb phrase ``sticking out" to ``springt aus" which means ``jump out" in German, while Text-Only NMT translates to ``streckt aus", which is correct. We find that VAG-NMT often makes mistakes when translating verbs. We think it is because the image vectors are pre-trained on an object classification task, which does not have any human action information.

\section{Conclusion and Future Work}
We proposed a visual grounding attention structure to take advantage of the visual information to perform machine translation. The visual attention mechanism and visual context grounding module help to integrate the visual content into the sequence-to-sequence model, which leads to better translation quality compared to the model with only text information. We achieved state-of-the-art results on the Multi30K and Ambiguous COCO dataset. We also proposed a new product dataset, IKEA, to simulate a real-world online product description translation challenge.

In the future, we will continue exploring different methods to ground the visual context into the translation model, such as learning a multimodal shared space across image, source language text, as well as target language text. We also want to change the visual pre-training model from an image classification dataset to other datasets that have both objects and actions, to further improve translation performance. 
\mingyang{
\section*{Acknowledge}
We would like to thank Daniel Boyla for providing insightful discussions to help with this research. We also want to thank Chunting Zhou and Ozan Caglayan for suggestions on machine translation model implementation. This work was supported in part by NSF CAREER IIS-1751206.
}
\bibliography{emnlp2018}
\bibliographystyle{acl_natbib_nourl}

\iffalse
\noindent {\bf Preparing References:} \\

Include your own bib file like this:
{\small\verb|\bibliographystyle{acl_natbib_nourl}|
\verb|\bibliography{emnlp2018}|}

\begin{thebibliography}{30}
\expandafter\ifx\csname natexlab\endcsname\relax\def\natexlab#1{#1}\fi

\bibitem[{Bahdanau et~al.(2014)Bahdanau, Cho, and Bengio}]{ref2}
Dzmitry Bahdanau, Kyunghyun Cho, and Yoshua Bengio. 2014.
\newblock Neural machine translation by jointly learning to align and
  translate.
\newblock \emph{CoRR}, abs/1409.0473.

\bibitem[{Caglayan et~al.(2017)Caglayan, Aransa, Bardet,
  Garc{\'{\i}}a{-}Mart{\'{\i}}nez, Bougares, Barrault, Masana, Herranz, and
  van~de Weijer}]{ref15}
Ozan Caglayan, Walid Aransa, Adrien Bardet, Mercedes
  Garc{\'{\i}}a{-}Mart{\'{\i}}nez, Fethi Bougares, Lo{\"{\i}}c Barrault, Marc
  Masana, Luis Herranz, and Joost van~de Weijer. 2017.
\newblock {LIUM-CVC} submissions for {WMT17} multimodal translation task.
\newblock \emph{CoRR}, abs/1707.04481.

\bibitem[{Calixto et~al.(2017{\natexlab{a}})Calixto, Liu, and Campbell}]{ref9}
Iacer Calixto, Qun Liu, and Nick Campbell. 2017{\natexlab{a}}.
\newblock Doubly-attentive decoder for multi-modal neural machine translation.
\newblock \emph{CoRR}, abs/1702.01287.

\bibitem[{Calixto et~al.(2017{\natexlab{b}})Calixto, Liu, and Campbell}]{ref11}
Iacer Calixto, Qun Liu, and Nick Campbell. 2017{\natexlab{b}}.
\newblock Incorporating global visual features into attention-based neural
  machine translation.
\newblock \emph{CoRR}, abs/1701.06521.

\bibitem[{Calixto et~al.(2017{\natexlab{c}})Calixto, Liu, and Campbell}]{ref43}
Iacer Calixto, Qun Liu, and Nick Campbell. 2017{\natexlab{c}}.
\newblock Multilingual multi-modal embeddings for natural language processing.
\newblock \emph{CoRR}, abs/1702.01101.

\bibitem[{Cho et~al.(2014)Cho, van Merrienboer, G{\"{u}}l{\c{c}}ehre, Bougares,
  Schwenk, and Bengio}]{ref46}
Kyunghyun Cho, Bart van Merrienboer, {\c{C}}aglar G{\"{u}}l{\c{c}}ehre, Fethi
  Bougares, Holger Schwenk, and Yoshua Bengio. 2014.
\newblock Learning phrase representations using {RNN} encoder-decoder for
  statistical machine translation.
\newblock \emph{CoRR}, abs/1406.1078.

\bibitem[{Chrupala et~al.(2015)Chrupala, K{\'{a}}d{\'{a}}r, and
  Alishahi}]{ref27}
Grzegorz Chrupala, {\'{A}}kos K{\'{a}}d{\'{a}}r, and Afra Alishahi. 2015.
\newblock Learning language through pictures.
\newblock \emph{CoRR}, abs/1506.03694.

\bibitem[{Denkowski and Lavie(2014)}]{ref30}
Michael Denkowski and Alon Lavie. 2014.
\newblock Meteor universal: Language specific translation evaluation for any
  target language.
\newblock In \emph{In Proceedings of the Ninth Workshop on Statistical Machine
  Translation}.

\bibitem[{Elliott et~al.(2017)Elliott, Frank, Barrault, Bougares, and
  Specia}]{ref3}
Desmond Elliott, Stella Frank, Lo{\"{\i}}c Barrault, Fethi Bougares, and Lucia
  Specia. 2017.
\newblock Findings of the second shared task on multimodal machine translation
  and multilingual image description.
\newblock \emph{CoRR}, abs/1710.07177.

\bibitem[{Elliott et~al.(2016)Elliott, Frank, Sima'an, and Specia}]{ref1}
Desmond Elliott, Stella Frank, Khalil Sima'an, and Lucia Specia. 2016.
\newblock Multi30k: Multilingual english-german image descriptions.
\newblock \emph{CoRR}, abs/1605.00459.

\bibitem[{Elliott and K{\'{a}}d{\'{a}}r(2017)}]{ref16}
Desmond Elliott and {\'{A}}kos K{\'{a}}d{\'{a}}r. 2017.
\newblock Imagination improves multimodal translation.
\newblock \emph{CoRR}, abs/1705.04350.

\bibitem[{Gella et~al.(2017)Gella, Sennrich, Keller, and Lapata}]{ref42}
Spandana Gella, Rico Sennrich, Frank Keller, and Mirella Lapata. 2017.
\newblock Image pivoting for learning multilingual multimodal representations.
\newblock \emph{CoRR}, abs/1707.07601.

\bibitem[{He et~al.(2015{\natexlab{a}})He, Zhang, Ren, and Sun}]{ref35}
Kaiming He, Xiangyu Zhang, Shaoqing Ren, and Jian Sun. 2015{\natexlab{a}}.
\newblock Deep residual learning for image recognition.
\newblock \emph{CoRR}, abs/1512.03385.

\bibitem[{He et~al.(2015{\natexlab{b}})He, Zhang, Ren, and Sun}]{ref45}
Kaiming He, Xiangyu Zhang, Shaoqing Ren, and Jian Sun. 2015{\natexlab{b}}.
\newblock Delving deep into rectifiers: Surpassing human-level performance on
  imagenet classification.
\newblock \emph{CoRR}, abs/1502.01852.

\bibitem[{Helcl and Libovick{\'{y}}(2017)}]{ref10}
Jindrich Helcl and Jindrich Libovick{\'{y}}. 2017.
\newblock {CUNI} system for the {WMT17} multimodal translation task.
\newblock \emph{CoRR}, abs/1707.04550.

\bibitem[{Huang et~al.(2016)Huang, Liu, Shiang, Oh, and Dyer}]{ref13}
Po-Yao Huang, Frederick Liu, Sz-Rung Shiang, Jean Oh, and Chris Dyer. 2016.
\newblock Attention-based multimodal neural machine translation.
\newblock In \emph{WMT}.

\bibitem[{Karpathy and Li(2014)}]{ref41}
Andrej Karpathy and Fei{-}Fei Li. 2014.
\newblock Deep visual-semantic alignments for generating image descriptions.
\newblock \emph{CoRR}, abs/1412.2306.

\bibitem[{Kingma and Ba(2014)}]{ref32}
Diederik~P. Kingma and Jimmy Ba. 2014.
\newblock Adam: {A} method for stochastic optimization.
\newblock \emph{CoRR}, abs/1412.6980.

\bibitem[{Kiros et~al.(2014)Kiros, Salakhutdinov, and Zemel}]{ref26}
Ryan Kiros, Ruslan Salakhutdinov, and Richard~S. Zemel. 2014.
\newblock Unifying visual-semantic embeddings with multimodal neural language
  models.
\newblock \emph{CoRR}, abs/1411.2539.

\bibitem[{Lin et~al.(2014)Lin, Maire, Belongie, Bourdev, Girshick, Hays,
  Perona, Ramanan, Doll{\'{a}}r, and Zitnick}]{ref37}
Tsung{-}Yi Lin, Michael Maire, Serge~J. Belongie, Lubomir~D. Bourdev, Ross~B.
  Girshick, James Hays, Pietro Perona, Deva Ramanan, Piotr Doll{\'{a}}r, and
  C.~Lawrence Zitnick. 2014.
\newblock Microsoft {COCO:} common objects in context.
\newblock \emph{CoRR}, abs/1405.0312.

\bibitem[{Ma et~al.(2017)Ma, Li, Zhao, and Huang}]{ref12}
Mingbo Ma, Dapeng Li, Kai Zhao, and Liang Huang. 2017.
\newblock {OSU} multimodal machine translation system report.
\newblock \emph{CoRR}, abs/1710.02718.

\bibitem[{Papineni et~al.(2002)Papineni, Roukos, Ward, and Zhu}]{ref31}
Kishore Papineni, Salim Roukos, Todd Ward, and Wei-Jing Zhu. 2002.
\newblock Bleu: A method for automatic evaluation of machine translation.
\newblock In \emph{Proceedings of the 40th Annual Meeting on Association for
  Computational Linguistics}, ACL '02, pages 311--318, Stroudsburg, PA, USA.
  Association for Computational Linguistics.

\bibitem[{Pascanu et~al.(2012)Pascanu, Mikolov, and Bengio}]{ref33}
Razvan Pascanu, Tomas Mikolov, and Yoshua Bengio. 2012.
\newblock Understanding the exploding gradient problem.
\newblock \emph{CoRR}, abs/1211.5063.

\bibitem[{Ren et~al.(2015)Ren, He, Girshick, and Sun}]{ref14}
Shaoqing Ren, Kaiming He, Ross Girshick, and Jian Sun. 2015.
\newblock Faster r-cnn: Towards real-time object detection with region proposal
  networks.
\newblock In \emph{Proceedings of the 28th International Conference on Neural
  Information Processing Systems - Volume 1}, NIPS'15, pages 91--99, Cambridge,
  MA, USA. MIT Press.

\bibitem[{Russakovsky et~al.(2015)Russakovsky, Deng, Su, Krause, Satheesh, Ma,
  Huang, Karpathy, Khosla, Bernstein, Berg, and Fei-Fei}]{ref36}
Olga Russakovsky, Jia Deng, Hao Su, Jonathan Krause, Sanjeev Satheesh, Sean Ma,
  Zhiheng Huang, Andrej Karpathy, Aditya Khosla, Michael Bernstein,
  Alexander~C. Berg, and Li~Fei-Fei. 2015.
\newblock {ImageNet Large Scale Visual Recognition Challenge}.
\newblock \emph{International Journal of Computer Vision (IJCV)},
  115(3):211--252.

\bibitem[{Schuster and Paliwal(1997)}]{ref25}
M.~Schuster and K.K. Paliwal. 1997.
\newblock Bidirectional recurrent neural networks.
\newblock \emph{Trans. Sig. Proc.}, 45(11):2673--2681.

\bibitem[{Sennrich et~al.(2017)Sennrich, Firat, Cho, Birch, Haddow, Hitschler,
  Junczys{-}Dowmunt, L{\"{a}}ubli, Barone, Mokry, and Nadejde}]{ref8}
Rico Sennrich, Orhan Firat, Kyunghyun Cho, Alexandra Birch, Barry Haddow,
  Julian Hitschler, Marcin Junczys{-}Dowmunt, Samuel L{\"{a}}ubli, Antonio
  Valerio~Miceli Barone, Jozef Mokry, and Maria Nadejde. 2017.
\newblock Nematus: a toolkit for neural machine translation.
\newblock \emph{CoRR}, abs/1703.04357.

\bibitem[{Sennrich et~al.(2015)Sennrich, Haddow, and Birch}]{ref44}
Rico Sennrich, Barry Haddow, and Alexandra Birch. 2015.
\newblock Neural machine translation of rare words with subword units.
\newblock \emph{CoRR}, abs/1508.07909.

\bibitem[{Young et~al.(2014)Young, Lai, Hodosh, and Hockenmaier}]{ref34}
Peter Young, Alice Lai, Micah Hodosh, and Julia Hockenmaier. 2014.
\newblock From image descriptions to visual denotations: New similarity metrics
  for semantic inference over event descriptions.
\newblock \emph{Transactions of the Association for Computational Linguistics},
  2:67--78.

\bibitem[{Zhang et~al.(2017)Zhang, Utiyama, Sumita, Neubig, and
  Nakamura}]{ref39}
Jingyi Zhang, Masao Utiyama, Eiichiro Sumita, Graham Neubig, and Satoshi
  Nakamura. 2017.
\newblock Nict-naist system for wmt17 multimodal translation task.
\newblock In \emph{WMT}.

\end{thebibliography}

Where \verb|emnlp2018| corresponds to the {\tt emnlp2018.bib} file.
\fi

\appendix
%\renewcommand{\thesection}{\Alph{section}}
% \documentclass[11pt,a4paper]{article}
% \usepackage[hyperref]{emnlp2018}
% \usepackage{times}
% \usepackage{latexsym}

% \usepackage{url}
% \usepackage{amsmath}
% \usepackage{graphicx}
% \usepackage[lofdepth,lotdepth]{subfig}
% \usepackage{multirow}
% \usepackage[utf8]{inputenc}
% \usepackage[T1]{fontenc}
% \aclfinalcopy

% \begin{document}
% \appendix
\section{IKEA Dataset Stats and Examples}
\label{appendix:IKEA}

\begin{figure}[ht]
    \centering
    \includegraphics[width=0.45\textwidth]{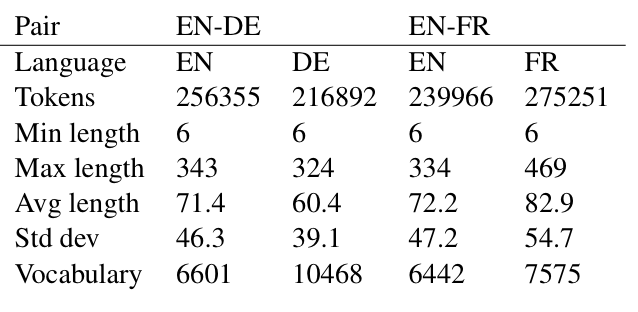}
    \caption{Statistic of the IKEA dataset}
    \label{fig:figure11}
\end{figure}

\begin{figure}[ht]
\centering
\subfloat[Subfigure 1 list of figures text][\textit{Product image}]{
\includegraphics[width=0.45\textwidth]{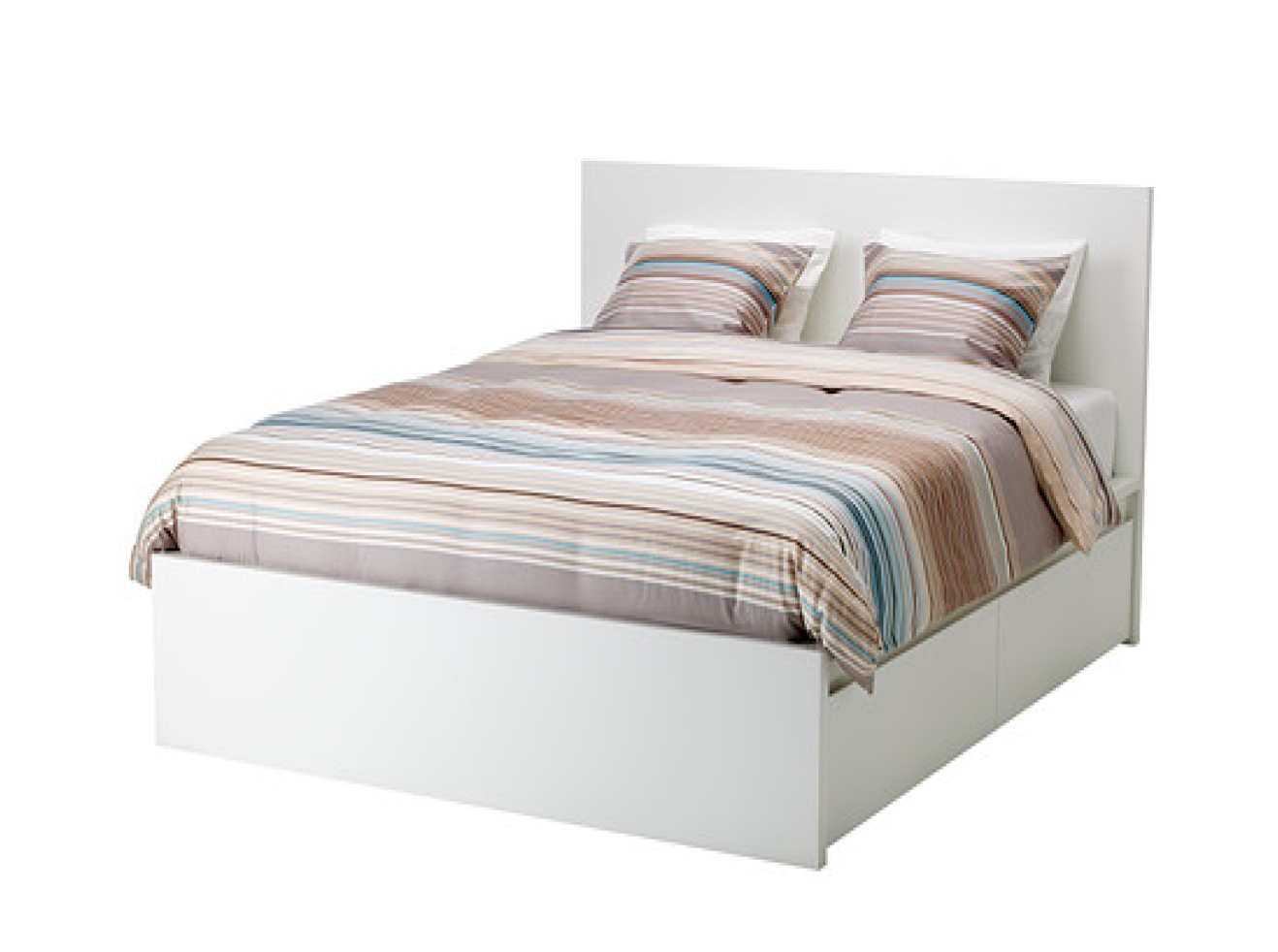}
\label{fig:subfig2}}
\qquad
\subfloat[Subfigure 2 list of figures text][ \textit{Source description}]{
\includegraphics[width=0.45\textwidth]{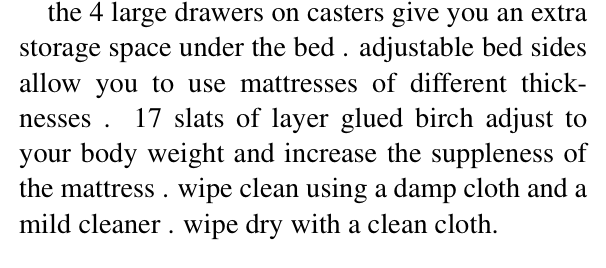}
\label{fig:subfig3}}
\qquad
\subfloat[Subfigure 3 list of figures text][ \textit{Target description in German}]{
\includegraphics[width=0.45\textwidth]{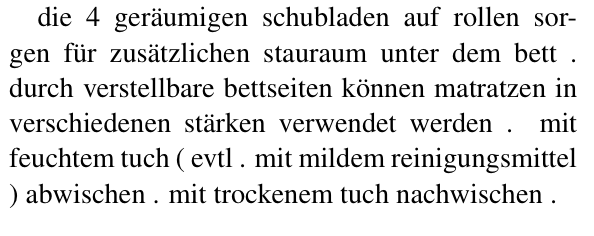}
\label{fig:subfig4}}

\caption{An example of product description and the corresponding translation in German from the \textbf{IKEA dataset}. Both descriptions provide an accurate caption for the commercial characteristics of the product in the image, but the details in the descriptions are different.}
\label{fig:figure10}
\end{figure}

We summarize the statistics of our IKEA dataset in Figure \ref{fig:figure11}, where we demonstrate the information about the number of tokens, the length of the product description and the vocabulary size. We provide one example of the IKEA dataset in Figure \ref{fig:figure10}.

\section{Hyperparameter Settings}
\label{appendix:Hyperparameters}
In this Appendix, we share details on the hyperparameter settings for our model and the training process. The word embedding size for both encoder and decoder are 256. The Encoder is a one-layer bidirectional recurrent neural network with Gated Recurrent Unit (GRU), which has a hidden size of 512. The decoder is a recurrent neural network with conditional GRU of the same hidden size. The visual representation is a 2048-dim vector extracted from the \texttt{pool5} layer of a pre-trained ResNet-50 network. The dimension of the shared visual-text semantic embedding space is 512. We set the decoder initialization weight value $\lambda$ to  0.5.   

During training, we use Adam \cite{ref32} to optimize our model with a learning rate of $4e-4$ for German Dataset and $1e-3$ for French dataset. The batch size is 32. The total gradient norm is clipped to 1 \cite{ref33}. Dropout is applied at the embedding layer in the encoder, context vectors extracted from the encoder and the output layer of the decoder. For Multi30K German dataset the dropout probabilities are $(0.3,0.5,0.5)$ and for Multi30K French dataset the dropout probabilities are $(0.2,0.4,0.4)$. For the  Multimodal shared space learning objective function, the margin size $\gamma$ is set to 0.1. The objective split weight $\alpha$ is set to 0.99. We initialize the weights of all the parameters with the method introduced in \cite{ref45}.

\section{Ablation Analysis on Visual-Text Attention}
We conducted an ablation test to further evaluate the effectiveness of our visual-text attention mechanism. We created two comparison experiments where we reduced the impact of visual-text attention with different design options. In the first experiment, we remove the visual-attention mechanism in our pipeline and simply use the mean of the encoder hidden states to learn the shared embedding space. In the second experiment, we initialize the decoder with just the mean of encoder hidden states without the weighted sum of encoder states using the learned visual-text attention scores. 

We run both experiments on Multi30K German Dataset five times and demonstrate the results in table  \ref{table:table_abtest}. As can be seen, the performance of the changed translation model is obviously worse than the full VAG-NMT in both experiments. This observation suggests that the visual-attention mechanism is critical in improving the translation performance. The model improvement comes from the attention mechanism influencing the model's objective function and decoder's initialization.

\begin{table}[h]
\small
\centering
\begin{tabular}{lll}
  & \multicolumn{2}{c}{English $\rightarrow$ German} \\
  \hline
  Method & BLEU & METEOR\\
  \hline
  -attention in shared embedding  & $30.5 \pm 0.6$ & $51.7 \pm 0.4$\\
  -attention in initialization  & $30.8 \pm  0.8$ & $51.9 \pm 0.5$\\
  VAG-NMT & $\boldsymbol{31.6 \pm 0.6}$ & $\boldsymbol{52.2 \pm 0.3}$ \\ 
\end{tabular}
\caption{Ablation analysis on visual-text attention mechanism in the Multi30K German dataset.}
\label{table:table_abtest}
\end{table}

\end{document}